\documentclass[10pt,letterpaper]{article}
\usepackage[latin1]{inputenc}
\usepackage{amsmath}
\usepackage{amsfonts}
\usepackage{amssymb}
\usepackage{amsthm}
\usepackage{makeidx}
\usepackage{multirow}

\usepackage{subfigure}

\usepackage{caption}

\usepackage{proof}
\usepackage{amsthm}

\usepackage{chngpage}

\usepackage[numbers]{natbib}

\usepackage[
	bookmarks=true, 
	colorlinks=true,
	linkcolor=blue, 
	pdftitle={Policy Gradient Methods for Off-policy Control},
    pdfauthor={Lucas Lehnert (lucas.lehnert@mail.mcgill.ca), Doina Precup (dprecup@cs.mcgill.ca)},
	]{hyperref}

\usepackage{todonotes}

\usepackage{algorithm,algorithmic}

\usepackage{tikz}
\usetikzlibrary{shapes,arrows,positioning}
\PassOptionsToPackage{usenames}{xcolor}
\definecolor{grey}{RGB}{229,229,229}
\definecolor{medium-grey}{RGB}{152,152,152}
\definecolor{dark-grey}{RGB}{122,122,122}

\usepackage{graphicx}
\usepackage{wrapfig}

\usepackage{epstopdf}

\theoremstyle{plain}

\theoremstyle{definition}

\author{Lucas Lehnert\\School of Computer Science\\McGill University\\\texttt{lucas.lehnert@mail.mcgill.ca} \and Doina Precup\\School of Computer Science\\McGill University\\\texttt{dprecup@cs.mcgill.ca}}
\title{Policy Gradient Methods for Off-policy Control}
\date{}

\begin{document}

\maketitle

\begin{abstract}
Off-policy learning refers to the problem of learning the value function of a way of behaving, or policy, while following a different policy. 
Gradient-based off-policy learning algorithms, such as GTD and TDC/GQ~\cite{Sutton09fastgradient}, converge even when using function approximation and incremental updates. 
However, they have been developed for the case of a fixed behavior policy. In control problems, one would like to adapt the behavior policy over time to become more greedy with respect to the existing value function. 
In this paper, we present the first gradient-based learning algorithms for this problem, which rely on the framework of policy gradient in order to modify the behavior policy. 
We present derivations of the algorithms, a convergence theorem, and empirical evidence showing that they compare favorably to existing approaches.
\end{abstract}

\section{Introduction}

One fundamental concept in Reinforcement Learning (RL) is Temporal Difference (TD) learning introduced by~\citet{Sutton1988}.
In TD-learning, methods such as TD(0) are used for policy evaluation where one tries to learn the value of a given state under a fixed policy.
The extension to the control case is called Q-learning where the value function is defined on state-action pairs.
The control policy is then computed from these action values.
One of the first Q-learning algorithms was proposed by~\citet{Watkins1992aa} which simultaneously searches and evaluates a policy by varying its action value estimates.
\citeauthor{Watkins1992aa}'s Q-learning algorithm is an off-policy algorithm as the policy that is searched and evaluated is strictly greedy with respect to the current action values, but for control the agent uses a $\varepsilon$-greedy policy.
This facilitates exploration as the agent is allowed to make a random move with $\varepsilon$ probability to obtain representative samples and facilitate the search for a policy that generates high rewards.

Recently, gradient-based off-policy learning algorithms were introduced such as GTD~\cite{SuttonGTD} and TDC~\cite{Sutton09fastgradient} which are also proven to be convergent under off-policy learning with linear value function approximation.
The extension to Q-learning, GQ($\lambda$)~\cite{SuttonGQ}, is also convergent under off-policy learning but only if the control policy is fixed. 
For the control case this is not sufficient as the agent has to explore its environment to be able to search and find a good policy.
The reason why convergence cannot be guaranteed is that a non-stationary policy causes drift in the distribution from which transition samples are generated.
While this drift is necessary for the agent to find a good policy, it can also cause oscillations in the value function estimates and the algorithm to not converge.
SARSA also suffers from this problem and is only guaranteed to converge to a sub-space of policies~\cite{Gordon01SARSARegion,Gordon96chatteringin}.
Within this sub-space the value function estimates may oscillate indefinitely.

In this paper we present a new gradient-based TD-learning algorithm that is similar to GQ but also incorporates policy gradients to correct for the drift in the distribution from which transitions are sampled.
Similar to the policy gradient framework~\cite{Sutton00policygradient} we directly analyze the interaction between the policy gradient and the distribution from which transitions are sampled.
As a result, our algorithm iterates over the sequence Markov Chains induced by the variation in the value function estimates and therefore policies.
This makes our algorithm similar to policy iteration methods such as~\cite{Perkins02PolicyIteration}.
However, rather than evaluating and then improving the policy in consecutive steps, our method simultaneously improves and evaluates the current policy.

\section{Q-learning with Policy Gradients}

We consider an MDP $\mathcal{M} = \langle \mathcal{S}, \mathcal{A}, t, r, \gamma \rangle$ where $\mathcal{S}$ is a finite state space and $\mathcal{A}$ is a finite action space.
The transition function $t : \mathcal{S} \times \mathcal{A} \times \mathcal{S} \to ( 0, 1 )$ is stochastic, the reward function is defined as $r : \mathcal{S} \times \mathcal{A} \to \mathbb{R}$, and the discount factor $\gamma \in (0,1)$.
As in~\cite{Sutton09fastgradient,SuttonGQ} we consider the linear function approximation case with a basis function $\phi : \mathcal{S} \times \mathcal{A} \to \mathbb{R}^k$ and define the state-action value function as
\begin{equation}
Q_\theta(s,a) = \theta^\top \phi(s,a) \approx Q(s,a) = \mathbb{E} \left[ \sum_{t=0}^\infty \gamma^t r_{t+1} \middle| s_0 = s, a_0 = a \right].
\end{equation} 
Let $Q_\theta = \Phi \theta \in \mathbb{R}^{|\mathcal{S} \times \mathcal{A}|}$ be the vector of all state-action values and similarly $R \in \mathbb{R}^{|\mathcal{S} \times \mathcal{A}|}$ be the vector of all rewards.
We are assuming that the MDP is ergodic and that a limit distribution $d_{s,a} = \lim_{t \to \infty} \mathbb{P} \{ s_t = s, a_t = a \}$ exists.
Letting $D$ be a diagonal matrix with the limit distribution on its diagonal we define the norm $|| \mathbf{v} ||_D^2 = \mathbf{v}^\top D \mathbf{v}$.
The Mean Squared Projected Bellman Error introduced by~\cite{Sutton09fastgradient} is
\begin{equation}
\text{MSPBE}(\theta) = || Q_\theta - \Pi T_\theta Q_\theta ||_D^2,
\end{equation}
where $\Pi = \Phi (\Phi D \Phi)^{-1} \Phi^\top D$ is the projection matrix and the Bellman operator applied to the action value function is defined as
\begin{equation}
T_\theta Q_\theta \overset{\text{def}}{=} R + \gamma P_\theta Q_\theta.
\end{equation}
Our approach differs to TDC and GQ in that we view the Bellman operator and the stationary distribution over state-action pairs as parametric in the value function parameter $\theta$.
For the stationary distribution we assume that 
\begin{equation}
d_{s,a} = d_s \pi_\theta (a | s) = \left[ \lim_{t \to \infty} \mathbb{P} \{ s_t = s \} \right] \pi_\theta (a | s) .
\end{equation}
This changes the way derive the gradient of the MSPBE as we assume additional dependencies on the parameter vector $\theta$ through the action selection probabilities $\pi_\theta (a|s)$.

\subsection{Gradient Derivation}

To obtain the gradient of the MSPBE objective,~\cite{Sutton09fastgradient} have shown
\begin{equation*}
\text{MSPBE}(\theta) = \left( \Phi^\top D (T_\theta Q_\theta - Q_\theta)  \right)^\top (\Phi^\top D \Phi)^{-1} \left( \Phi^\top D (T_\theta Q_\theta - Q_\theta) \right).
\end{equation*}
To simplify the gradient calculation we assume $\theta = [ \theta_1,...,\theta_n ]^\top$ and compute the partial dervative with respect to $\theta_i$, which we denote with $\partial_i$:
\begin{align*}
&\partial_i \text{MSPBE}(\theta) = \\
&=\partial_i \left[ \left( \Phi^\top D (T^\pi_\theta Q_\theta - Q_\theta)  \right)^\top (\Phi^\top D \Phi)^{-1} \left( \Phi^\top D (T^\pi_\theta Q_\theta - Q_\theta) \right) \right] \\
&=2 \partial_i \left[ \left( \Phi^\top D (T^\pi_\theta Q_\theta - Q_\theta)  \right)^\top \right] (\Phi^\top D \Phi)^{-1} \left( \Phi^\top D (T^\pi_\theta Q_\theta - Q_\theta) \right) \\
&\hspace{0.5in} + \left( \Phi^\top D (T^\pi_\theta Q_\theta - Q_\theta)  \right)^\top \partial_i \left[ (\Phi^\top D \Phi)^{-1} \right] \left( \Phi^\top D (T^\pi_\theta Q_\theta - Q_\theta) \right).
\end{align*}
For the derivative of the inverse feature covariance we have
\begin{align*}
\partial_i \left[ (\Phi^\top D \Phi)^{-1} \right] &= - (\Phi^\top D \Phi)^{-1} \partial_i (\Phi^\top D \Phi) (\Phi^\top D \Phi)^{-1} \\
&= - (\Phi^\top D \Phi)^{-1} (\Phi^\top \partial_i D \Phi) (\Phi^\top D \Phi)^{-1}.
\end{align*}
Plugging this back into the gradient above we obtain
\begin{align*}
&\partial_i \text{MSPBE}(\theta) = \\
&=2  \left( \Phi^\top \partial_i D (T^\pi_\theta Q_\theta - Q_\theta) + \Phi^\top D \partial_i (T^\pi_\theta Q_\theta - Q_\theta)  \right)^\top (\Phi^\top D \Phi)^{-1} \left( \Phi^\top D (T^\pi_\theta Q_\theta - Q_\theta) \right)  \\
&\hspace{0.5in} - \left( \Phi^\top D (T^\pi_\theta Q_\theta - Q_\theta)  \right)^\top (\Phi^\top D \Phi)^{-1} (\Phi^\top \partial_i D \Phi) (\Phi^\top D \Phi)^{-1} \left( \Phi^\top D (T^\pi_\theta Q_\theta - Q_\theta) \right).
\end{align*}
For the partial derivative on the Bellman error we have
\begin{align*}
\partial_i [T_\theta Q_\theta - Q_\theta] &= \partial_i [ R + \gamma P_\theta \Phi \theta - \Phi \theta ] \\
&= \gamma \partial_i P_\theta \Phi \theta + \gamma P_\theta \partial_i [ \Phi \theta ] - \partial_i [ \Phi \theta ] \\
&= \gamma \partial_i P_\theta \Phi \theta + \gamma P_\theta \Phi_{:,i} - \Phi_{:,i} ,
\end{align*}
where $\Phi_{:,i}$ is the $i$th column of $\Phi$.
Plugging this back into the MSPBE gradient we have
\begin{align}
&\partial_i \text{MSPBE}(\theta) \nonumber \\
&=2  \left( \Phi^\top \partial_i D (T_\theta Q_\theta - Q_\theta) + \Phi^\top D (\gamma \partial_i P_\theta \Phi \theta + \gamma P_\theta \Phi_{:,i} - \Phi_{:,i})  \right)^\top (\Phi^\top D \Phi)^{-1} \left( \Phi^\top D (T_\theta Q_\theta - Q_\theta) \right) \nonumber \\
&\hspace{0.5in} - \left( \Phi^\top D (T_\theta Q_\theta - Q_\theta)  \right)^\top (\Phi^\top D \Phi)^{-1} (\Phi^\top \partial_i D \Phi) (\Phi^\top D \Phi)^{-1} \left( \Phi^\top D (T_\theta Q_\theta - Q_\theta) \right) \nonumber \\
&=2  \left( \Phi^\top \partial_i D (T_\theta Q_\theta - Q_\theta) \right)^\top (\Phi^\top D \Phi)^{-1} \left( \Phi^\top D (T_\theta Q_\theta - Q_\theta) \right) \nonumber \\
&\hspace{0.5in} + 2 \left( \Phi^\top D (\gamma \partial_i P_\theta \Phi \theta + \gamma P_\theta \Phi_{:,i} - \Phi_{:,i})  \right)^\top (\Phi^\top D \Phi)^{-1} \left( \Phi^\top D (T_\theta Q_\theta - Q_\theta) \right) \nonumber \\
&\hspace{0.5in} - \left( \Phi^\top D (T_\theta Q_\theta - Q_\theta)  \right)^\top (\Phi^\top D \Phi)^{-1} (\Phi^\top \partial_i D \Phi) (\Phi^\top D \Phi)^{-1} \left( \Phi^\top D (T_\theta Q_\theta - Q_\theta) \right) . \label{eq:mspbe-partial-matrix}
\end{align}

\subsection{Sampling the Gradient}

To derive a stochastic gradient descend algorithm we rewrite~\eqref{eq:mspbe-partial-matrix} as expectations.
Let
\begin{equation*}
w = (\Phi^\top D \Phi)^{-1} \left( \Phi^\top D (T_\theta Q_\theta - Q_\theta) \right) = \mathbb{E} \left[ \phi \phi^\top \right]^{-1} \mathbb{E} \left[ \delta \phi \right],
\end{equation*}
where $\phi=\phi(s,a)$ with $s,a \sim d_{s,a}$ and the TD-error being
\begin{equation*}
\delta = r(s,a) + \gamma \sum_{a'} \theta^\top \phi' - \theta^\top \phi
\end{equation*}
with $\phi' = \phi_{s',a'}$ and $\mathbb{P} \{ s' | s,a \} =  t(s,a,s')$.
This simplifies the partial derivative to 
\begin{equation}
\partial_i \text{MSPBE}(\theta) =  2 \left[ \left( \Phi^\top \partial_i D (T_\theta Q_\theta - Q_\theta) \right)^\top + \left( \Phi^\top D (\gamma \partial_i P_\theta \Phi \theta + \gamma P_\theta \Phi_{:,i} - \Phi_{:,i})  \right)^\top  \right] w - w^\top (\Phi^\top \partial_i D \Phi) w .
\end{equation}
For the first matrix term we have
\begin{align*}
\Phi^\top \partial_i D (T_\theta Q_\theta - Q_\theta) &= \Phi^\top \partial_i D (R + \gamma P_\theta Q_\theta - Q_\theta) \\
&= \Phi^\top \text{diag} \left\{ \partial_i d_{s,a} \right\}_{s,a} (R + \gamma P_\theta Q_\theta - Q_\theta) \\
&= \Phi^\top \text{diag} \left\{ d_s \partial_i \pi_\theta(a | s) \right\}_{s,a} (R + \gamma P_\theta Q_\theta - Q_\theta) \\
&= \begin{bmatrix} d_{s_1} \partial_i \pi_\theta(a_1 | s_1) \phi_{s_1,a_1} & \cdots & d_{s_n} \partial_i \pi_\theta(a_m | s_n) \phi_{s_n,a_m} \end{bmatrix} \begin{bmatrix} \delta_{s_1,a_1} \\ \vdots \\ \delta_{s_n,a_m} \end{bmatrix} \\
&= \sum_{s,a} d_{s} \partial_i \pi_\theta(a | s) \phi^\top \delta \\
&= \mathbb{E} \left[ \frac{\partial_i \pi_\theta}{\pi_\theta} \delta \phi^\top \right],
\end{align*}
where the expectation is over $s,a \sim d_{s,a}$, $\partial_i \pi_\theta / \pi_\theta = \partial_i \pi_\theta(a|s) / \pi_\theta(a|s)$, and $\mathbb{P} \{ s' | s,a \} =  t(s,a,s')$ for the TD-error.
For the second matrix term we denote the $i$th component of $\phi$ as $\phi^i$. 
Expanding this term we have
\begin{align*}
& \Phi^\top D (\gamma \partial_i P_\theta \Phi \theta + \gamma P_\theta \Phi_{:,i} - \Phi_{:,i}) \\
&= \Phi^\top D \left( \gamma \begin{bmatrix} \sum_{s',a'} t(s_1,a_1,s') \partial_i \pi_\theta(a'|s') \theta^\top \phi' \\ \vdots \\  \sum_{s',a'} t(s_n,a_m,s') \partial_i  \pi_\theta(a'|s') \theta^\top \phi' \end{bmatrix} + \gamma \begin{bmatrix} \sum_{s',a'} t(s_1,a_1,s') \pi_\theta(a'|s') \phi'^i \\ \vdots \\  \sum_{s',a'} t(s_n,a_m,s') \pi_\theta(a'|s') \phi'^i \end{bmatrix} - \begin{bmatrix} \phi^i_{s_1,a_1} \\ \vdots \\ \phi^i_{s_n,a_m} \end{bmatrix} \right) \\
&= \Phi^\top D \left( \begin{bmatrix} \gamma \sum_{s',a'} t(s_1,a_1,s') \partial_i \pi_\theta(a'|s') \theta^\top \phi' + \gamma \sum_{s',a'} t(s_1,a_1,s') \pi_\theta(a'|s') \phi'^i - \phi^i_{s_1,a_1} \\ \vdots \\ \gamma \sum_{s',a'} t(s_n,a_m,s') \partial_i  \pi_\theta(a'|s') \theta^\top \phi' + \gamma \sum_{s',a'} t(s_n,a_m,s') \pi_\theta(a'|s') \phi'^i - \phi^i_{s_n,a_m} \end{bmatrix} \right) \\
&= \sum_{s,a} d_{s,a} \phi^\top  \left( \gamma \sum_{s',a'} t(s,a,s') \partial_i \pi_\theta(a'|s') \theta^\top \phi' + \gamma \sum_{s',a'} t(s,a,s') \pi_\theta(a'|s') \phi'^i - \phi^i \right) \\
&= \mathbb{E} \left[ \left( \gamma \mathbb{E} \left[ \frac{\partial_i \pi_\theta'}{\pi_\theta'} \theta^\top \phi' \right] + \gamma \mathbb{E} \left[ \theta^\top \phi'^i \right] - \phi^i \right) \phi^\top  \right] ,
\end{align*}
where $\partial_i \pi_\theta' / \pi_\theta' = \partial_i \pi_\theta(a'|s') / \pi_\theta(a'|s')$.
For the third term we obtain 
\begin{align*}
w^\top (\Phi^\top \partial_i D \Phi) w &= w^\top \left( \begin{bmatrix} \phi_{s_1,a_1} & \cdots & \phi_{s_n,a_m} \end{bmatrix} \text{diag} \left\{ d_s \partial_i \pi_\theta(a | s) \right\}_{s,a} \begin{bmatrix} \phi_{s_1,a_1} \\ \vdots \\ \phi_{s_n,a_m} \end{bmatrix} \right) w \\
&= w^\top \left( \sum_{s,a} d_s \partial_i \pi_\theta(a | s) \phi \phi^\top  \right) w \\
&= w^\top \mathbb{E} \left[ \frac{\partial_i \pi_\theta}{\pi_\theta} \phi \phi^\top  \right] w.
\end{align*}
Assembling the MSPBE gradient then gives
\begin{align}
&-\frac{1}{2} \nabla_\theta \text{MSPBE}(\theta) \nonumber \\
&= - \left\{ \mathbb{E} \left[ \frac{\nabla_\theta \pi_\theta}{\pi_\theta} \delta \phi'^\top \right] + \mathbb{E} \left[ \left( \gamma \mathbb{E} \left[ \frac{\nabla_\theta \pi_\theta'}{\pi_\theta'} \theta^\top \phi' \right] + \gamma \mathbb{E} \left[  \phi' \right] - \phi \right) \phi^\top  \right] \right\} w + \frac{1}{2} w^\top \mathbb{E} \left[ \frac{\nabla_\theta \pi_\theta}{\pi_\theta} \phi \phi^\top  \right] w \nonumber  \\
&= - \mathbb{E} \left[ \frac{\nabla_\theta \pi_\theta}{\pi_\theta} \delta \phi^\top \right] w - \mathbb{E} \left[ \left( \gamma \mathbb{E} \left[ \frac{\nabla_\theta \pi_\theta'}{\pi_\theta'} \theta^\top \phi' \right] + \gamma \mathbb{E} \left[  \phi' \right] - \phi' \right) \phi^\top  \right] w + \frac{1}{2} w^\top \mathbb{E} \left[ \frac{\nabla_\theta \pi_\theta}{\pi_\theta} \phi \phi^\top  \right] w \nonumber  \\
&= \mathbb{E} \left[ \phi \phi^\top  \right] \underbrace{\mathbb{E} \left[ \phi \phi^\top \right]^{-1} \mathbb{E} \left[ \delta \phi \right]}_{=w}  - \gamma \mathbb{E} \left[  \phi' \phi^\top \right] w  - \mathbb{E} \left[ \frac{\nabla_\theta \pi_\theta}{\pi_\theta} \delta \phi^\top \right] w - \gamma \mathbb{E} \left[ \frac{\nabla_\theta \pi_\theta'}{\pi_\theta'} \theta^\top \phi' \phi^\top \right] w + \frac{1}{2} w^\top \mathbb{E} \left[ \frac{\nabla_\theta \pi_\theta}{\pi_\theta} \phi \phi^\top  \right] w \nonumber  \\
&= \mathbb{E} \left[ \delta \phi \right] - \gamma \mathbb{E} \left[  \phi' \phi^\top \right] w  - \mathbb{E} \left[ \frac{\nabla_\theta \pi_\theta}{\pi_\theta} \delta \phi^\top \right] w - \gamma \mathbb{E} \left[ \frac{\nabla_\theta \pi_\theta'}{\pi_\theta'} \theta^\top \phi' \phi^\top \right] w + \frac{1}{2} w^\top \mathbb{E} \left[ \frac{\nabla_\theta \pi_\theta}{\pi_\theta} \phi \phi^\top  \right] w \label{eq:mspbe-grad}
\end{align}
To derive an iterative algorithm we follow~\cite{SuttonGTD} and derive two timescale update rules to learn the parameter vector $\theta$ and approximate the auxiliary weight vector $w$ with 
\begin{equation}
w \leftarrow w + \beta (\delta - \phi^\top w) \phi.
\end{equation}
Sampling the gradient above then gives the update rule
\begin{equation}
\theta \leftarrow \theta + \alpha \left[ \delta \phi - \gamma \phi' \phi^\top w - \frac{\nabla_\theta \pi_\theta}{\pi_\theta} \delta \phi^\top w - \gamma  \frac{\nabla_\theta \pi_\theta'}{\pi_\theta'} (\theta^\top \phi') (\phi^\top w) + \frac{1}{2} \frac{\nabla_\theta \pi_\theta}{\pi_\theta} (w^\top \phi)^2 \right].
\end{equation}
Note that this update rule contains the standard TDC/GQ term plus correction terms that are in the direction of the policy gradient. 

Algorithm~\ref{alg:pgq} shows the resulting algorithm, which we call PGQ for Policy-Gradient Q-learning.
This algorithm uses linear function approximation and updates are done in $O(k)$, where $k$ is the number of basis functions used.
After making a transition, we do not want to sample the next action using the old parameter estimate and rather use the updated $\theta$ estimate. 
To do this we have to the calculate expected values $\overline{\phi}$ and $\overline{\phi}^\nabla$ analytically over the next possible actions.

\begin{algorithm}                 
\caption{Policy Gradient Q-learning}
\label{alg:pgq}     
\begin{algorithmic}
\renewcommand{\algorithmicrequire}{\textbf{Input:}}    
\REQUIRE A transition sample $(s, a, r, s')$.
\STATE $\rho \leftarrow \frac{\pi_\theta(a|s)}{b(a|s)}$
\STATE $\rho^\nabla \leftarrow \frac{\nabla_\theta \pi_\theta(a|s)}{b(a|s)}$
\STATE $\overline{\phi} \leftarrow \sum_{a'} \pi_\theta(a'|s') \phi(s',a')$
\STATE $\overline{\phi}^\nabla \leftarrow \sum_{a'} \nabla_\theta \pi_\theta(a'|s') \theta^\top \phi(s',a')$
\STATE $\delta \leftarrow r + \gamma \theta^\top \overline{\phi} - \theta^\top \phi$
\STATE $\theta \leftarrow \theta + \alpha \left( \delta \phi - \gamma \phi' (\phi^\top w) - \rho^\nabla \delta (\phi^\top w) - \gamma \overline{\phi}^\nabla (\phi^\top w) + \frac{1}{2} (w^\top \phi)^2 \right)$
\STATE $w \leftarrow w + \beta ( \delta - \phi^\top w ) \phi$
\end{algorithmic}
\end{algorithm}

\section{Baird Counter Example}

\begin{figure}
\centering

\subfigure[On-policy Case: The Boltzmann temperature was set to $\tau=0.4$ and the learning rates are $\alpha = 0.01$ and $\beta = 0.25$.]{\includegraphics[width=0.48\textwidth]{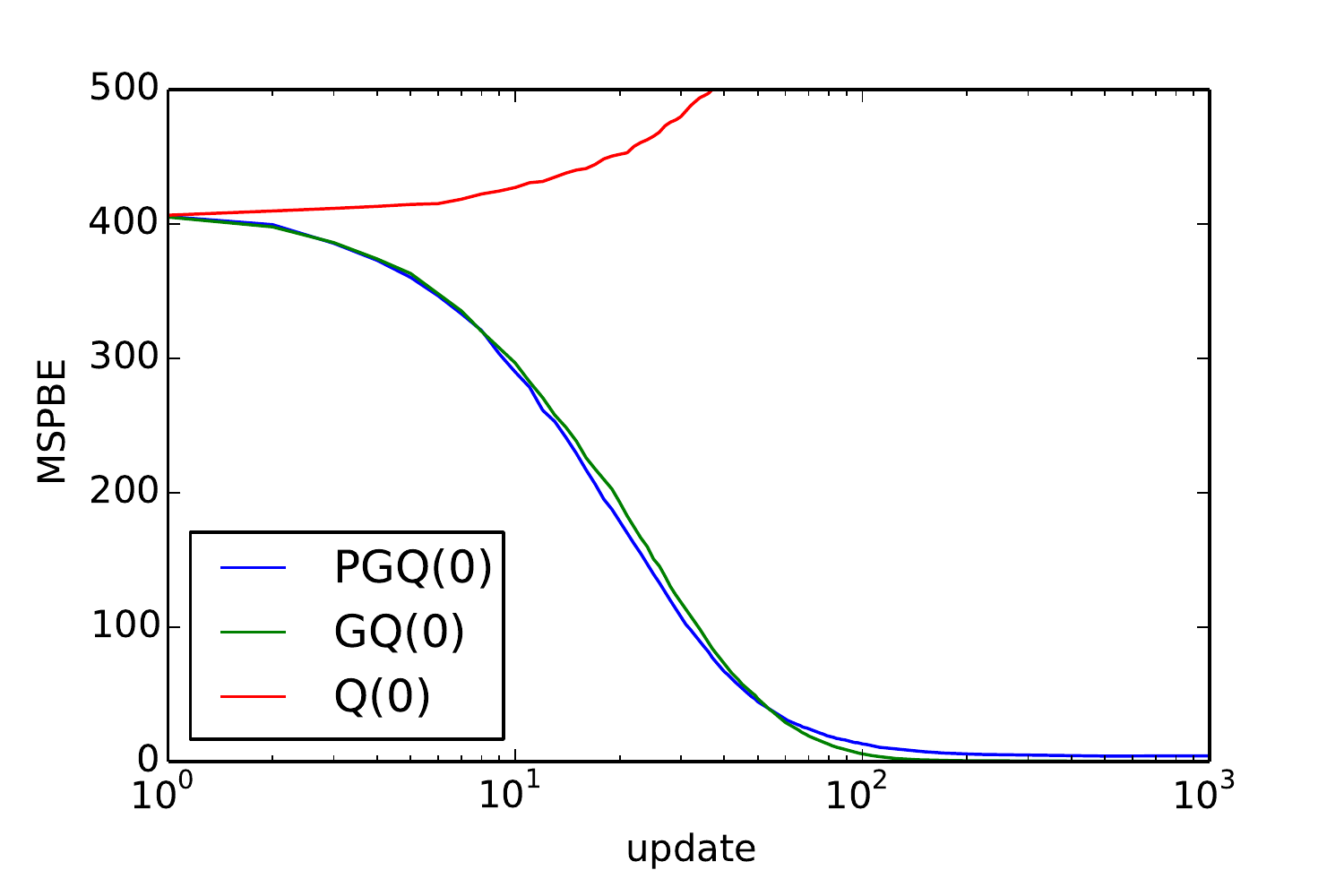}}
\subfigure[Off-policy Case: The Boltzmann temperature of the target policy was set to $0.4$ and the behaviour policy was set to $0.7$. The learning rates were set to $\alpha = 0.005$ and $\beta = 0.01$.]{\includegraphics[width=0.48\textwidth]{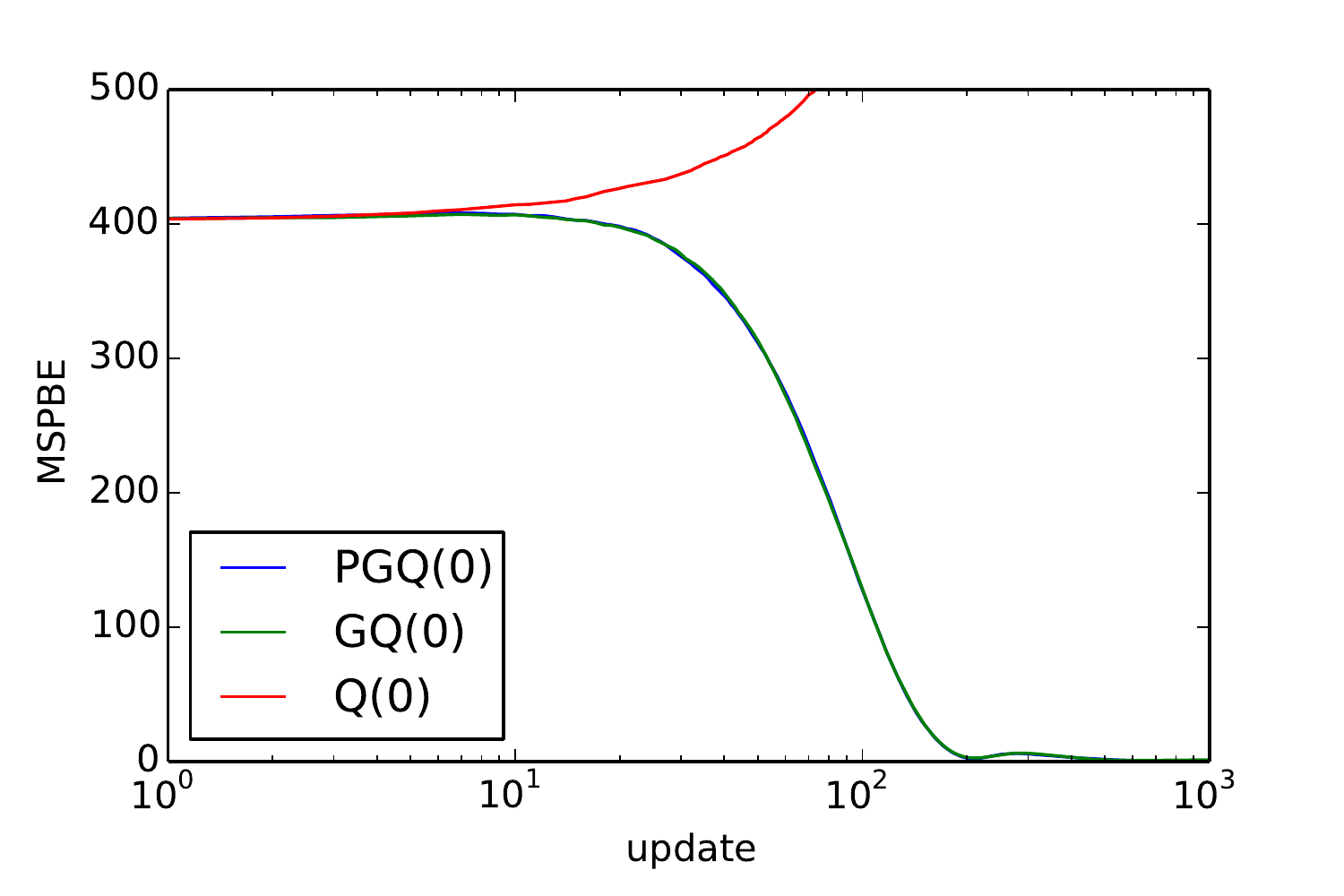}}

\caption{MSPBE for Q-learning, GQ, and PGQ on the Baird counter example with sampled updates.}
\label{fig:baird-sampled}
\end{figure}

We have tested our method on the "star" Baird counter example~\cite{Baird95} and compared it with Q-learning and GQ~\cite{SuttonGQ}.
For this 7 state version divergence of Q-learning is monotonic and GQ is known to converge~\cite{SuttonGreedyGQ}.
We initialize the parameter vector $\theta$ corresponding to the action that transitions to the 7th centre state with $(1,1,1,1,1,1,1,10)$ and the remaining parameter entries with 1.
The discount factor is set to $\gamma = 0.99$.
In our experiments we do not assume a hard-coded policy that ensures uniform exploration over state-action pairs but look at the control case where actions are selected using a Boltzmann policy where the probability of selecting a specific action is
\begin{equation}
\pi_\theta ( a | s ) = \frac{\text{exp} ( \theta^\top \phi(s,a) / \tau )}{ \sum_b  \text{exp} ( \theta^\top \phi(s,b) / \tau ) }.
\end{equation}
Updating was done either through sampling transitions $(s,a,s',r)$ according to hard coded distributions or either through simulating trajectories through the MDP.
For the sampled version we have sampled the state $s$ according to a uniform distribution over all 7 states, the action $a$ was sampled with probability $\pi_\theta (a|s)$ and the next state $s'$ was sampled according to the transition model.
Figure~\ref{fig:baird-sampled} shows the MSPBE error for the sampled update experiment.
Q-learning diverges monotonically and both GQ and PGQ converge to a zero MSPBE.

\begin{figure}
\centering

\subfigure{\includegraphics[width=0.48\textwidth]{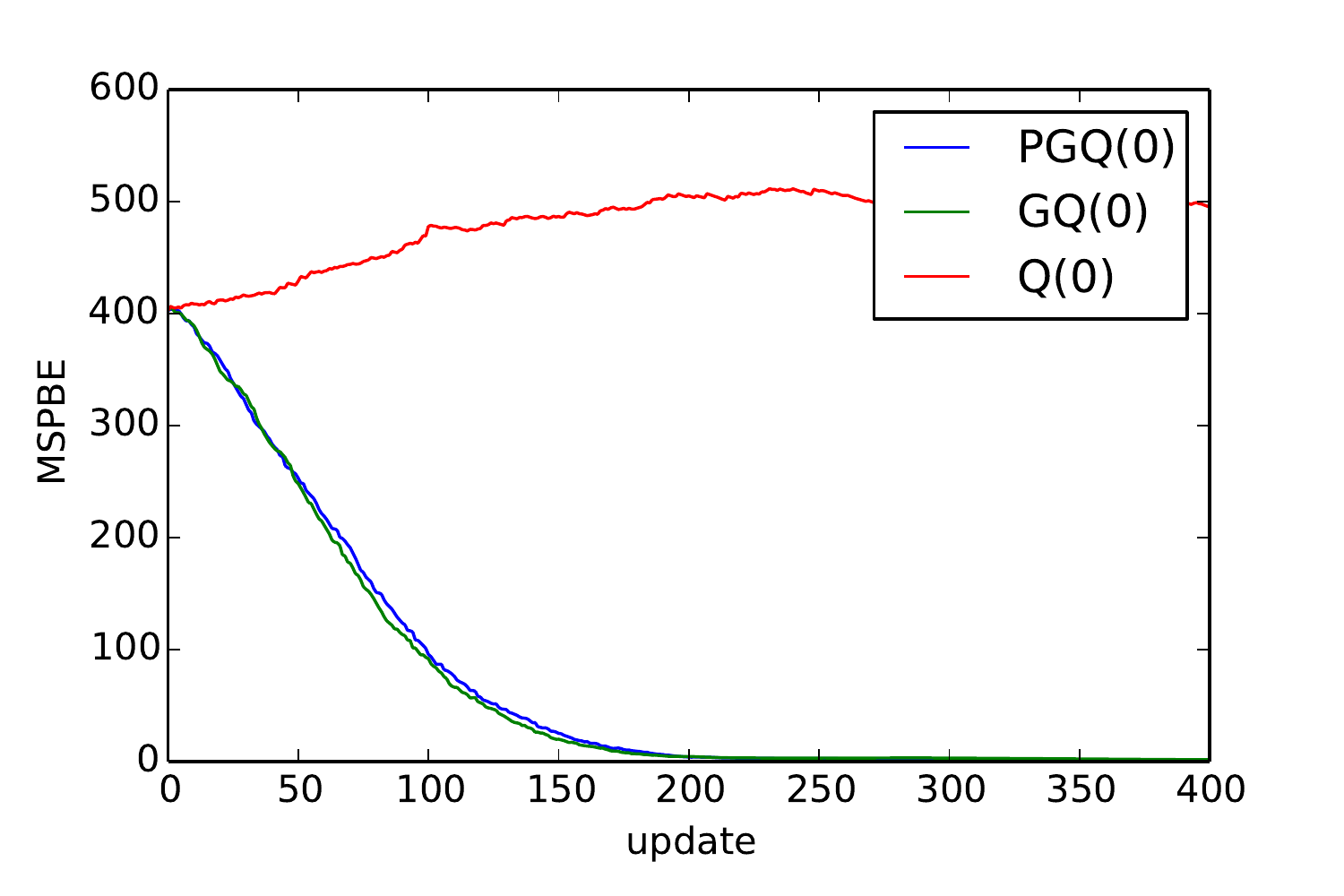}}
\subfigure{\includegraphics[width=0.48\textwidth]{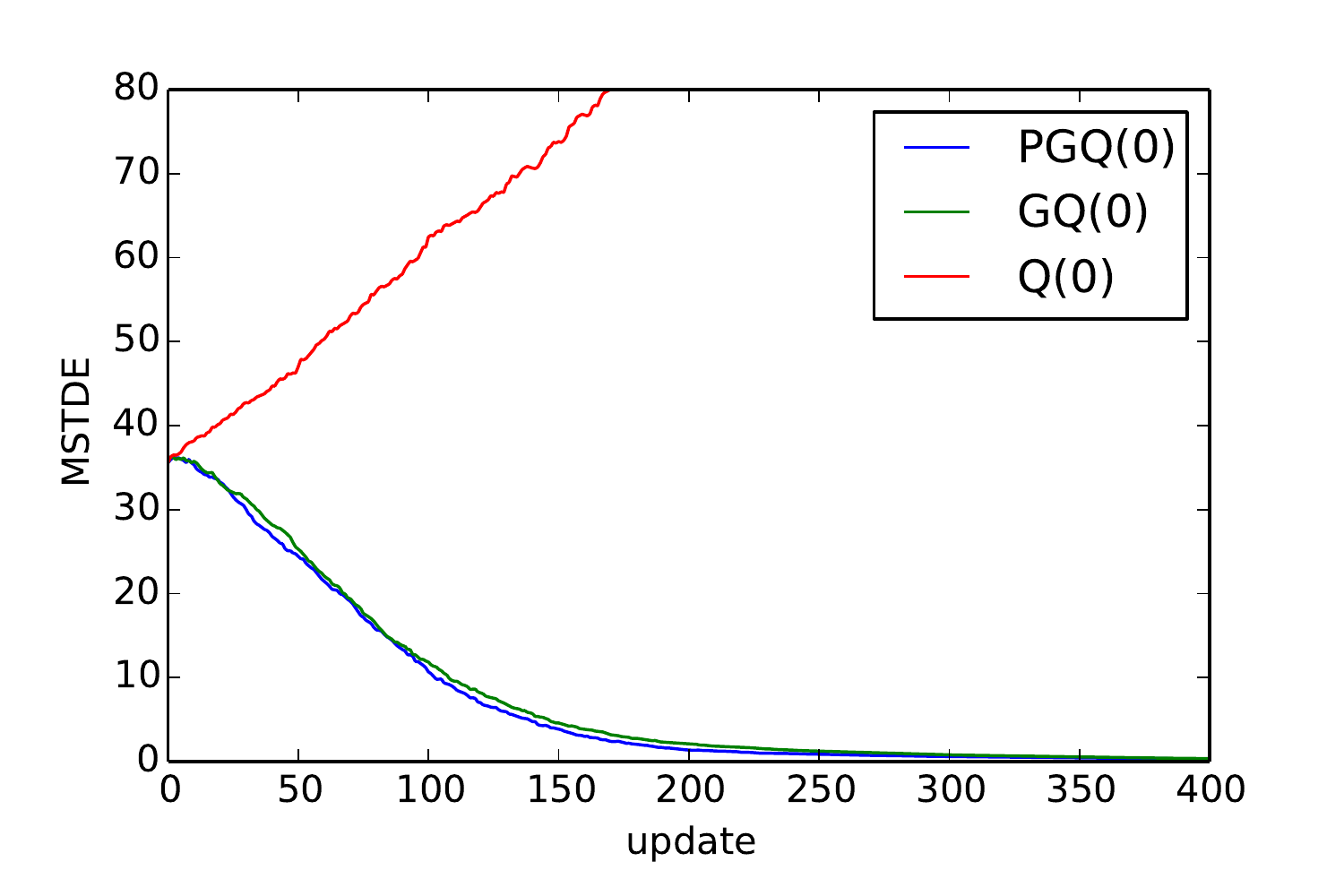}}

\caption{MSPBE and MSTDE for Q-learning, GQ, and PGQ on the Baird counter example with simulated trajectories. The learning rates were set to $\alpha = 0.0125$ and $\beta = 0.0421875$. The target policy temperature was set to 0.8 and the control policy temperature was set to 10.0. Each curve is the average of 10 repeated runs.}
\label{fig:baird-trajectory}
\end{figure}

For the trajectory based experiments we have sampled one of the seven start states uniformly and then executed transitions through the MDP.
While transitioning or updating the parameter vector we have measured the MSPBE using a uniform stationary distribution over states.
Figure~\ref{fig:baird-trajectory} shows the MSPBE and the Mean Squared TD-error (MSTDE) defined in~\cite{JMLRdann} of the parameter vector $\theta$ at each step of the simulation.

\section{Conclusion}

We have presented a new gradient based TD-learning algorithm that incorporates policy gradients.
The resulting algorithm is similar to GQ/TDC but also has a correction term in the direction of the gradient of the target policy.
Our analysis assumes a dependency of the Markov chain on the parameter vector $\theta$ through the target policy.
This allows our algorithm to correctly step over a sequence of different Markov chains and account for the drift in the distribution from which transition data is sampled due to changes in the parameter vector.


One next research direction is to extend this method to the non-linear function approximation case.
\citet{TDsmooth} present the first gradient based TD algorithm that converges in this case. 
One may able to draw on their results for our work.
For the derivation of our algorithm we only assumed the Bellman operator to be parametric in the parameter estimate, which lead to the additional policy gradient terms.
No further assumptions were made on the Bellman operator and the value function terms in the MSPBE objective, so in the non-linear function approximation case one would obtain gradients of the value function here.
However, one would have to analyze the projection operator in the MSPBE objective differently.

\bibliographystyle{plainnat}
\bibliography{pgtd_arXiv.bib}

\end{document}